\pdfoutput=1

\documentclass[11pt]{article}

\usepackage{EMNLP2022}

\usepackage{times}
\usepackage{latexsym}

\usepackage[T1]{fontenc}

\usepackage[utf8]{inputenc}

\usepackage{microtype}

\usepackage{inconsolata}

\usepackage{booktabs}
\usepackage{todonotes}
\usepackage{multirow}
\usepackage{graphicx,subcaption,ragged2e}

\makeatletter
\newcommand\footnoteref[1]{\protected@xdef\@thefnmark{\ref{#1}}\@footnotemark}
\makeatother

\title{Analyzing the Use of Influence Functions for Instance-Specific Data Filtering in Neural Machine Translation}

\author{Tsz Kin Lam\Thanks{\enspace Work done during an internship at Amazon.}\\ ICL, Heidelberg University \\ \texttt{lam@cl.uni-heidelberg.de}
         \And
         Eva Hasler \\ Amazon AI Translate \\ \texttt{ehasler@amazon.com}
         \And
         Felix Hieber \\ Amazon AI Translate \\ \texttt{fhieber@amazon.com}}

\begin{document}
\maketitle
\begin{abstract}
Customer feedback can be an important signal for improving commercial machine translation systems. 
One solution for fixing specific translation errors is to remove the related erroneous training instances followed by re-training of the machine translation system, which we refer to as instance-specific data filtering. Influence functions (IF) have been shown to be effective in finding such relevant training examples for classification tasks such as image classification, toxic speech detection and entailment task. Given a probing instance, IF find influential training examples by measuring the similarity of the probing instance with a set of training examples in gradient space. In this work, we examine the use of influence functions for Neural Machine Translation (NMT). We propose two effective extensions to a state of the art influence function  and demonstrate on the sub-problem of copied training examples that IF can be applied more generally than hand-crafted regular expressions.
\end{abstract}

\section{Introduction}

Neural Machine Translation (NMT) is the de facto standard for recent high-quality machine translation systems. NMT, however, requires abundant amount of bi-text for supervised training. One common approach to increase the amount of bi-text is via data augmentation \cite[\textit{inter alia}]{sennrich2015improving,edunov2018understanding,he2019revisiting}. Another approach is the use of web-crawled data \cite{banon2020paracrawl} but since crawled data is known to be notoriously noisy \cite{khayrallah2018impact,caswell2020language}, a plethora of data filtering techniques \cite[\textit{inter alia}]{junczys2018dual,wang2018denoising,ramirez2020bifixer} have been proposed for retaining a cleaner portion of the bi-text for training. 

While standard data filtering techniques aim to improve the quality of the overall training data without targeting the translation quality of specific instances, \textit{instance-specific data filtering} focuses on the improvement of translation quality toward a specific set of input sentences via removal of the related training data. In commercial MT, this selected set of sentences can be the problematic translations reported by customers. One simple approach of instance-specific data filtering in NMT is manual filtering. In manual filtering, human annotators identify translation errors on sentences reported by customer and designs filtering scheme, e.g., regular expressions to search related training examples for removal from the training set. 

In this work, we attempt to apply a more automatable technique called influence functions (IF) which is shown to be effective on image classification \cite{koh2017understanding}, and certain NLP tasks such as sentiment analysis, entailment and toxic speech detection \cite{han2020explaining,guo2020fastif}. Given a probing example, influence functions (IF) search for the influential training examples by measuring the similarity of the probing example with a set of training examples in gradient space. 
\citet{schioppa2021scaling} use a low-rank approximation of the Hessian to speed up the computation of IF and apply the idea of self-influence to NMT. However, self-influence measures if a training instance is an outlier rather than its similarity with another instance. \citet{tracingKnowledge} question the back-tracing ability of IF on the fact-tracing task. They compare IF with heuristics used in Information Retrieval and attribute the worse performance of IF to a problem called \textit{saturation}. Compared to fact-tracing, the target sides of machine translation can be more diverse which complicates the application of IF. 

We apply an effective type of IF called \textit{TracIn} \cite{pruthi2020estimating} to NMT for instance-specific data filtering and analyze its behaviour by constructing synthetic training examples containing simulated translation errors. In particular, we find that

\begin{itemize}
    \item the gradient similarity, also called the influence\footnote{In this work, we use gradient similarity or influence interchangeably to denote the result of IF. Be aware that TracIn is also one type of IF.}, is highly sensitive to the network component.
    \item vanilla IF may not be sufficient to achieve good retrieval performance. We proposed two contrastive methods to further improve the performance. 
    \item training examples consisting of copied source sentences have similar gradients even when they are lexically different. This indicates that the use of influence functions can go beyond what can be achieved with regular expressions. 
    \item an effective automation of the instance-specific data filtering remains challenging.
\end{itemize}

To the best of our knowledge, we are the first to investigate applying IF for instance-specific data filtering to NMT.

\section{Method}
\paragraph{Influence functions}
IF is a technique from robust statistics \cite[\textit{inter alia}]{hampel1974influence,cook1982residuals}. It aims to trace a model's predictions back to the most responsible training examples without repeated re-training of the model, aka Leave-One-Out. \citet{koh2017understanding} extend this idea from robust statistics to deep neural network that requires only the gradient of the loss functions $L$ and Hessian-vector products so that the influence $\mathcal{I}(z, z')$ of two examples $z$ and $z'$ is approximated as
\begin{equation}
    \mathcal{I}(z, z') \approx \nabla_{\theta}L(z')^{T}H^{-1}_{\hat{\theta}}\nabla_{\theta}L(z)
\end{equation}
where $\hat{\theta}$ is the model parameters at optimum and $H_{\hat{\theta}} = \frac{1}{n}\sum_{i=1}^{n}\nabla^{2}_{\theta}L(\theta)$ is the Hessian of the model parameters at $\hat{\theta}$. Given $n$ number of training instances and $p$ number of model parameters, the inverse of Hessian has a complexity of $\mathcal{O}(np^{2} + p^{3})$ which is expensive to compute for deep neural network. There are several proposed methods to speed up the computation of IF, e.g., by computing on a training subset selected by KNN-search \cite{guo2020fastif},  by approximating the Hessian with \textsc{LISSA} \cite{agarwal2017second}, by computing on a subset of model parameters \cite{koh2017understanding}, or by replacing the Hessian with some other procedures \cite{pruthi2020estimating}. In this work, we focus on TracIn which is shown to be better than some other variations \cite{han2020fortifying,schioppa2021scaling} in terms of retrieval performance.

TracIn, denoted by $\mathcal{I}\textsubscript{TracIn}(z, z')$, replaces the computationally costly Hessian matrix with an identity matrix. The remained gradient dot product, or called the gradient similarity, is instead computed over $C$ number of checkpoints, followed by averaging: 
\begin{equation}\label{eqt:TracIn}
    \mathcal{I}\textsubscript{TracIn}(z, z') = \frac{1}{C}\sum_{i=1}^{C}\nabla_{\theta}L(z')^{T}\nabla_{\theta}L(z)
\end{equation}
In NMT, given the same source sentence, the magnitude of the gradient in general is positively correlated to the length of the target sentence. In order to reduce the effect of the target length, we normalize equation \ref{eqt:TracIn} by the product of $\|\nabla_{\theta}L(z')\|$ and $\|\nabla_{\theta}L(z)\|$, or equivalently, we compute the cosine similarity of $\nabla_{\theta}L(z')$ and $\nabla_{\theta}L(z)$.

Given a probing instance $z'$ and its probing gradient $\nabla_{\theta}L(z')$, instances in the training set that yield a positive value of $\mathcal{I}\textsubscript{TracIn}(z, z')$ are called the positively influential training instances (+IFTrain) whereas those that yield a negative value of $\mathcal{I}\textsubscript{TracIn}(z, z')$ are called the negatively influential training instances (-IFTrain). Taking a gradient step on +IFTrain reduces the loss on the probing example while taking a gradient step on -IFTrain increases it. IF can be used for data filtering by removing the +IFTrain examples of low quality probing samples since their gradients have similar direction. Conversely, if the probing sample is of high quality, removing -IFTrain examples from the training data would be expected to increase translation quality w.r.t. the probing sample.

\begin{table*}[ht!]
\resizebox{\textwidth}{!}
{
\begin{tabular}{cl|cc|ccccc}
\toprule
& & \multicolumn{2}{c|}{Shared parameters} & \multicolumn{4}{c}{Non-shared parameters} \\
& Samples & $\nabla_{Full}$ & $\nabla_{Emb}$ & $\nabla_{srcEmb}$ & $\nabla_{trgEmb}$ & $\nabla_{output}$ & $\nabla_{concat}$ \\
\midrule
Probing & Noch kommt Volkswagen glimpflich durch. & 1 & 1 & 1 & 1 & 1 & 1 \\
  & Volkswagen gets off lightly. & & & & & \\
\midrule
1  & \textcolor{red}{Das £ 1,35 Mrd. teure Projekt soll bis} & 0.153 & 0.240 & 0.006 & 0.287 & 0.437 & 0.339 \\
  & \textcolor{red}{Mai 2017 fertiggestellt werden} & & & & &  \\
& Volkswagen gets off lightly. & & & & & \\
\midrule
2  & \textcolor{red}{Alle in Frage kommenden Produkte wurden} & 0.238 & 0.320 & 0.013 & 0.230 & 0.401 & 0.319 \\
  & \textcolor{red}{ aus dem Verkauf gezogen.} & & & & &  \\
  & Volkswagen gets off lightly. & & & & & \\
\midrule
3  & Noch kommt Volkswagen glimpflich durch. & -0.021 & -0.030 & -0.149 & -0.022 & -0.017 & -0.040 \\
& \textcolor{red}{In 2008, most malware programmes were} & & & & &  \\
& \textcolor{red}{still focused on sending out adverts.} & & & & &  \\
\midrule
4  & Noch kommt Volkswagen glimpflich durch. & -0.007 & -0.016 & -0.120 & -0.003 & 0.011 & -0.013 \\
  & \textcolor{red}{We've made a complete turnaround.} & & & & & \\
\midrule
5  & Noch kommt Volkswagen glimpflich durch. & 0.950 & 0.894 & 0.973 & 0.927 & 0.843 & 0.873  \\
  & Volkswagen gets off lightly\textcolor{red}{!} & & & & & \\
\midrule
6  & Noch kommt Volkswagen glimpflich durch\textcolor{red}{!} & 0.899 & 0.912 & 0.873 & 0.915 & 0.940 & 0.927 \\
  & Volkswagen gets off lightly. & & & & & \\
\bottomrule
\end{tabular}
}
\caption{Example showing the changes of influence by network components. Segments that are marked in red are perturbed from the probing example. $\nabla_{X}$ indicates the network components used in computing the influence, $\nabla_{concat}$ indicates the concatenation of $\nabla_{srcEmb}$, $\nabla_{trgEmb}$ and $\nabla_{output}$. }
\label{tab:sentivity_network}
\end{table*}

\section{Experimental Setting}
\paragraph{Model configuration and training}
We use Transformer BASE configuration as described in \citet{vaswani2017attention} with default setting and implementation in \textsc{fairseq}. We use a sentence-piece model to create subword units of size 32k. Unless otherwise specified, we pre-trained our NMT on Europarl-v7 data and News Commentary-v12 data in German-English direction from WMT17 for 100 epochs, about 112K updates, using Adam optimizerion training of 16-bit\footnote{We use 32-bit precision to compute the gradient similarity once the training is done.}. The effective mini-batch size is 4096 x 16 tokens and it takes a p3.16xlarge\footnote{\label{awsspecs}See https://aws.amazon.com/ec2/instance-types/ for details.} machine on AWS 6 hours for training. We evaluate the MT model on the newstest2017 test set with a checkpoint averaged over the 10-best checkpoints, measured by the validation loss on the newstest2014-2016 dev set. On the test set, our NMT model with non-shared parameters with the two word embeddings and the output layer scores 29.99 BLEU whereas the one with shared parameters scores 29.78 BLEU. We use beam search with beam size of 5 in decoding.

\paragraph{TracIn}
We select 5 checkpoints, i.e., at epoch 5, 8, 15, 30 and 100 for computing TracIn\footnote{It is tempting to just use the deployed checkpoint to compute the influence. As shown by Liang et al. 2017, however, the Hessian term in equation 1 captures more accurately the effect of model training than the dot product of the optimal checkpoint. In TracIn, the Hessian is approximated by the average over a set of checkpoints, and we follow their guidelines for checkpoints selection.}. We select checkpoints which have relatively large changes in the validation loss, i.e., usually in the earlier phrase of training, and include the last one to cover information at the end of the training. We compute the per-sample gradient with a batch size of 1 parallelized over multiple processes with several g4dn.2x\footnoteref{awsspecs} machines on AWS.

\section{Experimental results}
This section describes our findings on the properties of applying IF on NMT for instance-specific data filtering. 

\subsection{Sensitivity of gradient similarity to the network components}
In previous works, the influence, or called the gradient similarity, is usually computed with respect to a small part of the network parameters, especially the last or the last few layers (\citet{han2020explaining};\citet{barshan2020relatif}; \textit{inter alia}). In NMT, we found that the resulting influence is highly sensitive to the network components used in computing the gradients (or gradient component). 
For illustration, we construct a set of perturbed instances, compute its influence by different gradient components and observe their changes. The perturbed instances are not included during the NMT training. This independence between the NMT and the perturbed instances provides a simpler setting for checking how gradient components and the perturbed examples affect the influence.

Table~\ref{tab:sentivity_network} shows the gradient similarities of a probing example from newstest2017
with six artificially created instances. We use two NMT models, 1) trained with shared parameters between the two word embeddings and the output layer and 2) trained without parameter sharing, to compute the similarities. 


We notice that gradient similarity for the model with shared parameters is more strongly influenced by lexical matches on the target side, as shown by the larger magnitude of influence values for probing examples 1 and 2 with random source sides compared to probing examples 3 and 4 with random target sides. For non-shared parameters, we observe that the gradient w.r.t. the output layer ($\nabla_{output}$) has stronger response (0.437 and 0.401) to the probing instances with random source side whereas the gradient w.r.t. source embedding ($\nabla_{srcEmb}$) has stronger response (-0.149 and -0.120) to the instances with random target sides. On the same probing example, we repeat this random sampling of source and target sentences by using the other 3003 instances in the newstest2017 set. We find that the mean magnitude of $\nabla_{srcEmb}$ is 0.04 for random target whereas it is 0.004 for random source. In the case of $\nabla_{output}$, the mean magnitude for random target is 0.021 whereas it is 0.428 for random source. This indicates that $\nabla_{output}$ has a tendency of scoring sentence pairs higher when their target side overlaps with the target side of the probing instance and is less influenced by source-side overlap. This may be suboptimal for retrieving problematic training examples that are relevant to a given probing instance.

When using a gradient vector $\nabla_{concat}$ which is the concatenation of $\nabla_{srcEmb}$, $\nabla_{trgEmb}$ and $\nabla_{output}$, its similarity is dominated by $\nabla_{output}$ rather than equally shared between the three given that they have the same number of parameters. This may explain why, in the case of shared parameters, instances with random source side have higher similarities than those with random target side. 

Instance 5 and 6 are minor edits of the probing instance with changes to punctuation.
For instance 5, it is not easy to interpret the results for the model with shared parameters. However, in the non-shared parameter setting, we observe a higher similarity for $\nabla_{srcEmb}$ than for $\nabla_{trgEmb}$ and $\nabla_{output}$. This is more interpretable because the punctuation change is on the target side. For instance 6, the punctuation change is on the source side and we see a higher TracIn value for $\nabla_{output}$ than for $\nabla_{srcEmb}$ and $\nabla_{trgEmb}$. As before, the value of $\nabla_{concat}$ is more similar to the value of $\nabla_{output}$. Further examples can be found in Table \ref{tab:sensitivity_network_2} in the Appendix.

These qualitative results show that the choice of network component is crucial in computing the gradient similarity. As shown in the next experiment, this affects the retrieval of training examples. 

\subsection{Contrastive signal is crucial for better retrieval performance}
In this section, we try to illustrate how different gradient components affect the retrieval of the noisy instances with TracIn. We add control to the retrieval outcome by adding synthetic noisy training instances to the training data. In addition, we show that vanilla IF may not be sufficient to achieve good performance because the gradients are aggregated over all tokens in the target sentence. We thus propose two contrastive methods to sharpen the gradient signal.

\paragraph{Synthetic noisy examples} We use the error template \textit{X $\rightarrow$ Y} which stands for \textit{X is translated to Y} to construct synthetic noise examples for the training set . We created four simple error patterns: 1) \textit{August $\rightarrow$ January}, 2) \textit{Deutschland $\rightarrow$ Italy}, 3) \textit{Oktober $\rightarrow$ December} and 4) \textit{Türkei $\rightarrow$ New Zealand}.

\begin{table}[ht]
\resizebox{\columnwidth}{!}{
\begin{tabular}{lccc}
\toprule
\multirow{2}{2cm}{Error pattern} & \multicolumn{3}{c}{Number of instances} \\
 & {train} & {synthetic noisy}  & {probing} \\
\midrule
\textit{August} $\rightarrow$ \textit{January} & 8,017 & 925 & 9 \\
\midrule
\textit{Deutschland} $\rightarrow$ \textit{Italy} & 15,360 & 4,891 & 30 \\
\midrule
\textit{Oktober} $\rightarrow$ \textit{December} & 11,927 & 2,422 & 8 \\
\midrule
\textit{Türkei} $\rightarrow$ \textit{New Zealand} & 14,963 & 7,417 & 22 \\
\bottomrule
\end{tabular}
}
\caption{Number of instances per error pattern}
\label{tab:data_noisy}
\end{table}

In the training set, we replace the translation of the sentences containing the source pattern by the erroneous translation with a probability of 60\% so that the total number of training data is unchanged. We select these error patterns because translation errors of months and country names can easily result from noisy training examples and are therefore suitable to simulate real customer issues. In addition, there are related source sentences in the test set, i.e., newstest2017, which can be used as probing examples. In order to speed up the computation of IF, we extract a subset of training data containing the original pattern, the perturbed pattern and some randomly sampled training sentences. For example, in the error pattern \textit{Oktober $\rightarrow$ December}, the training subset contains sentences with \textit{Oktober}, \textit{Dezember}, \textit{October} and \textit{December} on either the source or target side together with some randomly sampled sentences. Table \ref{tab:data_noisy} gives the exact number of instances for each case. We follow the same training procedure as section 3 to pre-train a NMT model on the training corpus perturbed by the synthetic noises. 

\paragraph{Contrastive-IF} The gradient of a source-target pair in NMT involves complicated mapping between the source tokens and the target tokens. That is, the gradient vector does not just contain the information of the error pattern but also other context. In order to isolate the gradient of the error pattern from the aggregated signal, we propose two methods: 1) gradient masking and 2) gradient difference. Both methods leverage a cleaner translation either in the form of a gold-reference translation or a corrected hypothesis, i.e. the hypothesis with the error pattern corrected. We refer to them as \textit{Contrastive Influence Functions} (Contrastive-IF). 

\begin{table*}[ht]
\footnotesize
{
\begin{subtable}[h]{\textwidth}
\centering
\begin{tabular}{lcccc}
\toprule
\multirow{2}{*}{$\nabla$(Probing)} & \multirow{2}{*}{+/-} & \multicolumn{3}{c}{Precision} \\
  & & $\nabla_{srcEmb}$ & $\nabla_{output}$ & $\nabla_{Full}$ \\
\midrule
$\nabla$(HYP) & + & 0.846 &	0.720 &	0.503 \\
$\nabla$(REF) & - & 0.876 &	0.794 &	0.481 \\
$\nabla$(CorrHYP) & - & 0.930 &	0.905 &	0.531 \\ [2mm]
$\nabla$(HYP\textsubscript{Mask}) & + & 0.893 & 0.840 &	0.654 \\
$\nabla$(HYP\textsubscript{MaskExact}) & + & 0.957 & 0.910 & 0.862 \\
$\nabla$(CorrHYP\textsubscript{MaskExact})& - & 0.989 &	\bf 0.992 &	0.924 \\ [2mm]
$\nabla$(HYP) - $\nabla$(REF) & + & 0.930 &	0.856 &	0.584 \\
$\nabla$(HYP) - $\nabla$(CorrHYP) & + & \bf 1.000 &	0.971 &	\bf 0.987 \\
\bottomrule
\end{tabular}
\caption{Retrieval performance for top-1\% influential training examples}
\end{subtable}
\vspace*{5mm}
\newline
\begin{subtable}[h]{\textwidth}
\centering
\begin{tabular}{lcccc}
\toprule
\multirow{2}{*}{$\nabla$(Probing)} & \multirow{2}{*}{+/-} & \multicolumn{3}{c}{Precision} \\
  & & $\nabla_{srcEmb}$ & $\nabla_{output}$ & $\nabla_{Full}$ \\
\midrule
$\nabla$(HYP) & + & 0.765 & 0.644 &	0.442 \\
$\nabla$(REF) & - & 0.799 &	0.693 &	0.437 \\
$\nabla$(CorrHYP) & - & 0.844 &	0.781 &	0.455 \\ [2mm]
$\nabla$(HYP\textsubscript{Mask}) & + & 0.848 &	0.829 &	0.567 \\
$\nabla$(HYP\textsubscript{MaskExact})  & + & 0.936 & 0.904 & 0.825 \\ 
$\nabla$(CorrHYP\textsubscript{MaskExact}) & - & 0.962 & \bf 0.958 & 0.875 \\ [2mm]
$\nabla$(HYP) - $\nabla$(REF) & + & 0.855 &	0.764 &	0.515 \\
$\nabla$(HYP) - $\nabla$(CorrHYP) & + & \bf 0.986 &	0.935 & \bf 0.931 \\
\bottomrule
\end{tabular}
\caption{Retrieval performance for top-10\% influential training examples}
\end{subtable}
}
\caption{Retrieval performance measured in (macro) averaged precision over all error patterns. $\nabla(Probing)$ refers to the gradient with input ‘source-Probing’. HYP, REF and CorrHYP stands for hypothesis, reference and corrected hypothesis respectively. ``+'' (``-'') indicates that positively (negatively) influential training instances were retrieved. $\nabla_{X}$ indicates network components used in computing the gradient. We mark the best result per column in bold.}
\label{tab:avgPrecision}
\end{table*}

The idea of gradient masking (\textit{Mask}) is to apply a 0/1 token-level mask to the loss function so as to remove the contribution of irrelevant tokens from the gradient computation. We assign the mask based on which tokens differ between hypothesis and reference. If the 0-mask is applied everywhere except for the location of the error according to a corrected translation, we refer to it as \textit{MaskExact}.

We can use the difference between two hypotheses in a continuous fashion by simply subtracting their gradients. Specifically, we compute the difference of the gradient of a sentence $A$ and the gradient of a sentence $B$ as the probing gradient: $GD(A, B) = \nabla(A) - \nabla(B)$.
In this work, we use the hypothesis as $A$ and a cleaner translation as $B$ (either the reference or the corrected hypothesis) so that positively influential training instances w.r.t. to $GD(A,B)$ are the synthetic noisy training instances.

\paragraph{Results} Table \ref{tab:avgPrecision} shows the retrieval performance of vanilla IF, gradient masking and gradient difference where the gradient is computed w.r.t. to either the source embedding, output layer or the full model. We evaluate the performance with precision over the top-X\% influential training instances, i.e. the number of synthetic training instances successfully retrieved given top-X\% of the influential training samples. We combine results of the four error patterns  by (macro) averaging their precision. 

The first three rows show results for vanilla IF (TracIn) when either the hypothesis, the reference or a corrected hypothesis is used for probing the training data. 
Using $\nabla_{srcEmb}$ or $\nabla_{output}$ obtain substantially higher precision for each variant than using $\nabla_{Full}$, i.e., the gradient w.r.t. the entire model, which demonstrates the importance of the choice of gradient component(s) in vanilla-IF for retrieval performance. Using the corrected hypotheses to retrieve negatively-influential examples yields the best precision for both top-1\% and top-10\% of retrieved training examples.

We qualitatively examine the influential instances retrieved. By using the source-hypothesis pair as the probing instance, we find that instances retrieved via $\nabla_{output}$ have less similarity on the source side. In the first probing example, \textit{Januar $\rightarrow$ January} occurs more frequently in the ranking than \textit{August $\rightarrow$January}. In the second example, \textit{Italien $\rightarrow$ Italy} appears as the third influential training instance when using $\nabla_{output}$ whereas all top-3 influential instances obtained by $\nabla_{srcEmb}$ contain the desired error pattern of \textit{Deutschland $\rightarrow$ Italy}, see Table \ref{tab:qualitative_analysis} in the Appendix.

We find that both gradient masking, $\nabla$(HYP\textsubscript{Mask}),
and gradient difference, $\nabla$(HYP) $-$ $\nabla$(REF), perform better than the vanilla IF given the same gradient component. 
$\nabla$(HYP\textsubscript{Mask}) always outperforms the comparable vanilla IF variants $\nabla$(HYP) and $\nabla$(REF). If we can identify the exact location of the error pattern, with the probing gradient $\nabla$(HYP\textsubscript{MaskExact}) or $\nabla$(CorrHYP\textsubscript{MaskExact}), the precision can be further boosted and this is consistent for gradients $\nabla_{srcEmb}$, $\nabla_{output}$ and $\nabla_{Full}$. While the gradient difference variants do not always outperform the comparable masking variants for all $\nabla_{X}$, $\nabla$(HYP) $-$ $\nabla$(CorrHYP) yields the overall best result using $\nabla_{srcEmb}$.

An interesting finding is the improvement brought by the corrected hypothesis (CorrHYP). Applying vanilla-IF on it already achieves a precision of 0.930 under $\nabla_{srcEmb}$ considering the top-1\% influential instances. By applying \textit{MaskExact} or gradient difference on it, we achieve very high precisions of 0.989 and 1.0 under $\nabla_{srcEmb}$ considering the top-1\% influential training instances. One notable gain brought by the proposed approaches is that for $\nabla_{Full}$, the precision increases from 0.531 to around 0.987 for the $\nabla$(HYP) $-$ $\nabla$(CorrHYP) variant, bringing it on-par to the performance of $\nabla_{output}$. We include results for additional gradient components in Table~\ref{tab:avgPrecisionFull} in the Appendix.


\begin{table}[h]
\resizebox{\columnwidth}{!}
{
\begin{tabular}{lcccc}
\toprule
\multirow{2}{*}{$\nabla$(Probing)} & top-X\% influential & \multirow{2}{*}{+/-} & \multicolumn{2}{c}{Precision} \\
& training samples & & $\nabla_{Emb}$ & $\nabla_{Full}$ \\
\midrule
\multirow{2}{*}{$\nabla$(HYP)} & 1\% & \multirow{2}{*}{+} & 0.660	& 0.502 \\
& 10\% & & 0.596 & 0.444 \\
\midrule
\multirow{2}{*}{$\nabla$(CorrHYP)} & 1\% & \multirow{2}{*}{-} & 0.877 & 0.541 \\
& 10\% & & 0.746 & 0.463 \\
 \midrule
\multirow{2}{*}{$\nabla$(HYP) - $\nabla$(CorrHYP)} & 1\% & \multirow{2}{*}{+} & 0.891	& 0.691 \\
& 10\% & & 0.808 & 0.607 \\
\bottomrule
\end{tabular}
}
\caption{Retrieval performance measured in average precision across all error patterns for an NMT model with shared parameters between the word embeddings and the output layer.}
\label{tab:avgPrecision_shared}
\end{table}

\begin{table}[ht]
\setlength{\tabcolsep}{4pt}
\footnotesize
{
\begin{subtable}[h]{0.45\textwidth}
\centering
\begin{tabular}{lcccc}
\toprule
\multirow{2}{*}{$\nabla$(Probing)} & \multirow{2}{*}{+/-} & \multicolumn{3}{c}{Precision} \\
  & & $\nabla_{srcEmb}$ & $\nabla_{encoder}$ & $\nabla_{Full}$ \\
\midrule
$\nabla$(HYP) & + & 0.930 & 0.972 & 0.994 \\
$\nabla$(REF) & - & 0.525 & 0.452 & 0.548 \\
$\nabla$(HYP) - $\nabla$(REF) & + & 0.708 & 0.712 & 0.949 \\
\bottomrule
\end{tabular}
\caption{Retrieval performance for top-10\% influential training examples}
\end{subtable}
\vspace*{5mm}
\newline
\begin{subtable}[h]{0.45\textwidth}
\centering
\begin{tabular}{lcccc}
\toprule
\multirow{2}{*}{$\nabla$(Probing)} & \multirow{2}{*}{+/-} & \multicolumn{3}{c}{Precision} \\
  & & $\nabla_{srcEmb}$ & $\nabla_{encoder}$ & $\nabla_{Full}$ \\
\midrule
$\nabla$(HYP) & + & 0.888 & 0.932 & 0.986 \\
$\nabla$(REF) & - & 0.508 & 0.449 & 0.504 \\
$\nabla$(HYP) - $\nabla$(REF) & + & 0.670 & 0.647 & 0.895 \\
\bottomrule
\end{tabular}
\caption{Retrieval performance for top-20\% influential training examples}
\end{subtable}
}
\caption{Retrieval performance measured in averaged precision over the probing instances, on copied training instances. $\nabla(Probing)$ refers to the gradient with input ‘source-Probing’. HYP, REF stands for hypothesis, reference. ``+'' (``-'') indicates that positively (negatively) influential training instances were retrieved. $\nabla_{X}$ indicates the network components used in computing the gradient.}
\label{tab:avgPrecision_untranslatedSRC}
\end{table}

We also conducted a side experiment with a NMT model with shared parameters between the embeddings and the output layer. Similar to the case of a NMT model with non-shared parameters, gradient difference improves over the vanilla-IF when averaging precisions over all error patterns as shown in Table \ref{tab:avgPrecision_shared}.

To summarize, both our contrastive-IF variants improve retrieval performance regardless of the network component used in computing gradients and whether the NMT model has shared parameters. 

\begin{table*}[ht]
\centering
\footnotesize
{
\begin{tabular}{lcccc}
\toprule
Error pattern & \multicolumn{2}{c}{$\nabla$(HYP) - $\nabla$(CorrHYP)} & \multicolumn{2}{c}{$\nabla$(HYP)} \\
& $\nabla_{srcEmb}$ & $\nabla_{Full}$ & $\nabla_{srcEmb}$ & $\nabla_{Full}$ \\
\midrule
\textit{August} $\rightarrow$ \textit{January} & 0.399 $\pm$ 0.104 & 0.199 $\pm$ 0.041 & 0.059 $\pm$ 0.023 & 0.119 $\pm$ 0.042 \\
\textit{Oktober} $\rightarrow$ \textit{December} & 0.524 $\pm$ 0.192 & 0.397 $\pm$ 0.123 & 0.056 $\pm$ 0.028 & 0.143 $\pm$ 0.043 \\
\textit{Deutschland} $\rightarrow$ \textit{Italy} & 0.576 $\pm$ 0.126 & 0.428 $\pm$ 0.047 & 0.097 $\pm$ 0.061 & 0.135 $\pm$ 0.046 \\
\textit{Türkei} $\rightarrow$ \textit{New Zealand} & 0.527 $\pm$ 0.100 & 0.540 $\pm$ 0.118 & 0.080 $\pm$ 0.044 & 0.165 $\pm$ 0.051 \\
\end{tabular}
}
\caption{Statistics showing the mean and standard deviation of the largest influence per configuration. The large standard deviation of the maximum influence value for probing examples of the same error pattern shows the difficulty of defining a comparable filtering threshold across probing instances.}
\label{tab:avgMaxIF}
\end{table*}

\subsection{Copied source sentences have similar gradient signature}
Our initial motivation for applying influence functions to NMT was to arrive at a more automatable way of retrieving relevant training examples for reported translation problems. We were also hoping to generalize over what can be achieved by applying manually composed regular expressions which are limited to detecting lexical overlap. In this section, we focus on the latter and investigate whether Influence Functions can retrieve training examples that cause an undesired copy behaviour in the decoder. 

\paragraph{Experimental settings} On top-of the Europarl-v7 and News Commentary-v12 data, we append a set of  176,004 copied source sentences provided by \citet{khayrallah2018impact} to the training set. Following the training recipe in section 3, our NMT with non-shared parameters has a degradation of translation quality from 29.99 BLEU to 17.64 BLEU on the newstest2017 data, showing the detrimental effect of the untranslated target sides.

We select 40 probing instances from the newstest2017 data where their translation by the above NMT model is a copy of the source sentence. We again reduce the computation time by running TracIn over a training subset which contains the newly added noisy data, i.e., 176,004 instances and a set of randomly sampled training instances. This creates a training subset of 476,004 instances. 

\paragraph{Results}
Table \ref{tab:avgPrecision_untranslatedSRC} shows the retrieval performance on copied source sentences in the training subset with probing gradients of $\nabla(HYP)$, $\nabla(REF)$ and $\nabla(HYP)$ - $\nabla(REF)$ computed over source embedding ($\nabla_{srcEmb}$), the encoder ($\nabla_{encoder}$), or the entire model ($\nabla_{Full}$). We skip the masking strategy in this case since it would mask all target tokens, resulting in a loss of 0. Different from our results so far, the vanilla IF using only the hypothesis preforms better than using the reference for retrieval and better than the gradient difference variant for all network components. For example, when considering only the top-10\% influential training instances, the precision is 0.930 for $\nabla(HYP)$ with $\nabla_{srcEmb}$ and only 0.525 for $\nabla(REF)$. This may indicate that instances of copied source sentence have similar gradient signature despite their lexical difference (see Table \ref{tab:qualitative_analysis_untranslatedSRC} for some examples) and that the reference translation is less useful in this setting because it cannot provide a specific contrastive signal. 

A surprising finding in this setting is that using gradients computed over the entire network is better than the source embedding or the entire encoder. This is in contrast to the previous findings in the synthetic training instances. This possibly indicates that the copy mechanism is spread over the entire model or parts beyond the source embedding or the encoder.

\begin{table*}[ht]
\centering
\footnotesize
{
\begin{tabular}{lcccc}
\toprule
Error pattern & \multicolumn{2}{c}{$\nabla$(HYP) - $\nabla$(CorrHYP)} & \multicolumn{2}{c}{$\nabla$(HYP)} \\
& $\nabla_{srcEmb}$ & $\nabla_{Full}$ & $\nabla_{srcEmb}$ & $\nabla_{Full}$ \\
\midrule
\textit{August} $\rightarrow$ \textit{January} & 1.44 $\pm$ 0.50 & 3.33 $\pm$ 1.76 & 1.78 $\pm$ 1.55 & 1.44 $\pm$ 0.69 \\
\textit{Oktober} $\rightarrow$ \textit{December} & 2.25 $\pm$ 0.43 & 2.00 $\pm$ 0.00 & 2.88 $\pm$ 1.76 & 2.00 $\pm$ 1.58 \\
\textit{Deutschland} $\rightarrow$ \textit{Italy} & 1.00 $\pm$ 0.00 & 1.77 $\pm$ 0.62 & 1.67 $\pm$ 1.22 & 2.70 $\pm$ 2.62 \\
\textit{Turkei} $\rightarrow$ \textit{New Zealand} & 3.05 $\pm$ 1.46 & 1.32 $\pm$ 1.26 & 2.27 $\pm$ 2.09 & 2.32 $\pm$ 1.66 \\
\end{tabular}
}
\caption{Mean and standard deviation of the number of influential training instances to be removed per configuration, using the largest consecutive difference found in the ranking as clustering criterion.}
\label{tab:cluster}
\end{table*}

\subsection{An effective IF-based instance-specific data filtering is hard to automate}
Many data filtering algorithms require a threshold to decide which instances are to be filtered. This threshold can be a model score in an offline filtering algorithm \cite{junczys2018dual} or a dynamic formula that is changed according to the learning state of the model \cite{wang2018denoising}. In both cases, a desirable threshold should be effective as measured in the downstream model performance and be easily computed and generalized to other situations. In the case of IF-based instance-specific data filtering, we observe two properties in the ranking of the influence which makes the automation of the data filtering algorithm challenging. 

\paragraph{1: The range of influence varies across probing examples} Although the influence is bounded between $[-1, 1]$ because of the cosine similarity, the maximum magnitude of the influence for each probing example can still be very different. Table \ref{tab:avgMaxIF} shows the mean and standard deviation of the maximum influence value of positively influential training instances computed over probing examples of the same configuration. Firstly, the mean value is quite diverse across different gradient components, and across different probing gradients of the same error pattern. For example, the mean value of the error pattern \textit{August $\rightarrow$ January} computed with $\nabla_{srcEmb}$ is 0.399 or 0.059 depending on which probing gradient is used. Secondly, the standard deviation within each configuration is relatively large when compared to the corresponding mean value. For example, it is about 26\%, 36\%, 22\% and 19\% in the case of $\nabla_{srcEmb}$ using gradient difference as the probing gradient. This large standard deviation indicates the difficulty of setting an effective threshold for filtering even for probing examples with the same type of error pattern.

\begin{figure*}[ht]
\begin{subfigure}[t]{0.45\textwidth}
\includegraphics[width=\textwidth]{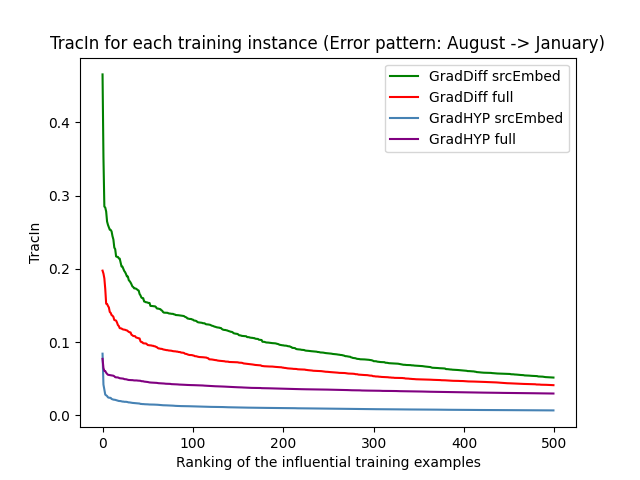}
\end{subfigure}\hspace{\fill}
\begin{subfigure}[t]{0.45\textwidth}
\includegraphics[width=\textwidth]{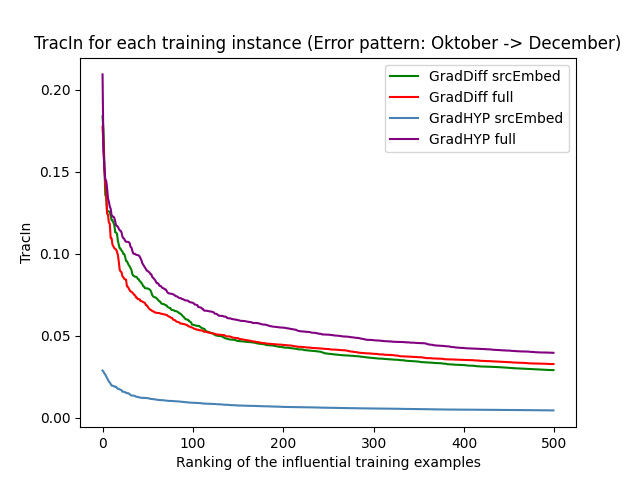}
\end{subfigure}

\begin{subfigure}[t]{0.45\textwidth}
\includegraphics[width=\textwidth]{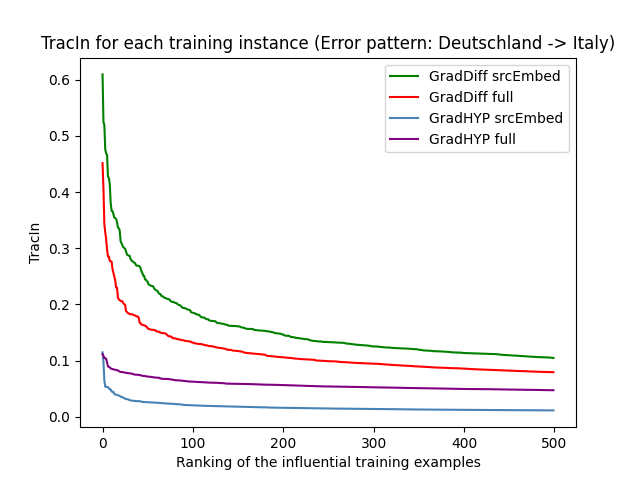}
\end{subfigure}\hspace{\fill}
\begin{subfigure}[t]{0.45\textwidth}
\includegraphics[width=\textwidth]{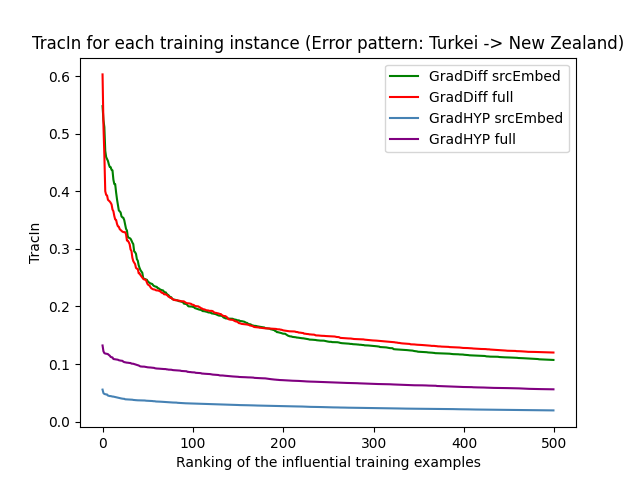}
\end{subfigure}
\caption{TracIn of the top-500 positively influential training examples. In each subfigure, we randomly select a probing example from each error pattern to compute its influence using gradient difference w.r.t. 1) source embedding (GradDiff srcEmbed), 2) entire model (GradDiff full) and using vanilla-IF with source-hypothesis as input w.r.t. 1) source embedding (GradHYP srcEmbed), 2) entire model (GradHYP full).}
\label{fig:IF_ranking_500}
\end{figure*}

\paragraph{2: The influence value drops abruptly at the top-of the ranking} Apart from a fixed threshold across different probing example, we also examine the possibility of automatically setting a threshold for each probing example. 

We first examine a simple clustering strategy by searching for the position where the consecutive difference is the largest in the ranking of influence. Table \ref{tab:cluster} shows the result of the mean and standard deviation of the number of most influential training instances to be removed per configuration. By considering only the largest consecutive difference, less than 5 training instances would be removed which is far less than the number of synthetic training instances.

We examine further by investigating the shape of the influence of the positively influential training instances in the ranking. Figure \ref{fig:IF_ranking_500} shows the influences, computed via TracIn, of the top-500 positively influential training instances per error pattern. For each error pattern, we randomly select a probing example to examine its influence under different gradient conditions. In all these cases, the influence drops sharply in the first few instances, especially in the case of vanilla IF, denoted by ``GradHYP'' in the figures. After the sharp drop, the influence becomes quite steady for the remaining instances. This steady behaviour holds even for instances of much lower rank, see Figure \ref{fig:IF_ranking_half} in the Appendix. The ``elbow'' occurs before the first 50 influential training instances, which includes only a tiny portion of the synthetic noisy training instances. 

 \paragraph{How about Top-K filtering?} In previous work, the authors use either Top-K or Top-X\% as the filtering threshold which is not realistic in the case of NMT where 1) there can be billions of training instances, and 2) the error types are more diverse than the prediction of wrong classes. In spite of the good retrieval performance demonstrated in the previous section, our results here show that an effective automation of the IF-based instance-specific data filtering for NMT remains a challenge.


\section{Conclusion}
We have analyzed the use of Influence Functions for NMT as instance-specific data filtering. By constructing synthetic instances, we found that 1) the gradient similarity is very sensitive to the selected network components, 2) vanilla Influence Functions are not sufficient for good retrieval performance, 3) our proposed contrastive-IF can boost the retrieval performance regardless of the gradient component or parameter sharing, 4) finding an effective automation of IF for instance-specific data filtering is difficult. This is because the proper choice of gradient component with respect to the type of error in the probing example is crucial for the effectiveness of Influence Functions.  
Despite the reported effectiveness for certain classification tasks in previous literature, our results show that applying IF to NMT poses some practical difficulties that we have not yet been able to solve. 

\section{Limitations}

In this work, we provided an analysis of using Influence Functions for Neural Machine Translation as instance-specific data filtering for the purpose of cost saving and finding a more generally applicable solution. Despite the reported success of some previous works in NLP/Vision-related classification tasks, we faced several challenges in applying Influence Functions to NMT. 
We are aware of the following limitations to our analysis:

\begin{itemize}
\item Our analysis focuses on TracIn rather than other influence functions because TracIn is reported to be very effective.
\item Our analysis is based on a fixed set of checkpoints, following the practice of previous works. The selection and the number of checkpoints used in TracIn are computationally costly hyper-parameters.
\item Our analysis focuses on major network components such as embeddings,  encoder and the output layer, excluding other possible combinations.
\item The scale of our experiments is limited, e.g., only the De-En language direction with 3M training instances and the synthetic examples are relatively simple. However, given such simple setting, we can already see the challenges of applying IF on NMT as
instance-specific data filtering or as an attribution/interpretable method.
\item The proposed contrastive IF requires a corrected translation, e.g., reference translation. 
\end{itemize}

We hope that our analysis can inspire further evaluation and modification of the technique. 

\section*{Acknowledgements}
We sincerely thank Ke Tran, Tobias Domhan, Bill Byrne and Sony Trenous for fruitful discussions and support.

\pagebreak

\bibliography{references}

\begin{thebibliography}{21}
\expandafter\ifx\csname natexlab\endcsname\relax\def\natexlab#1{#1}\fi

\bibitem[{Agarwal et~al.(2017)Agarwal, Bullins, and Hazan}]{agarwal2017second}
Naman Agarwal, Brian Bullins, and Elad Hazan. 2017.
\newblock Second-order stochastic optimization for machine learning in linear
  time.
\newblock \emph{The Journal of Machine Learning Research}, 18(1):4148--4187.

\bibitem[{Akyürek et~al.(2022)Akyürek, Bolukbasi, Liu, Xiong, Tenney,
  Andreas, and Guu}]{tracingKnowledge}
Ekin Akyürek, Tolga Bolukbasi, Frederick Liu, Binbin Xiong, Ian Tenney, Jacob
  Andreas, and Kelvin Guu. 2022.
\newblock Tracing knowledge in language models back to the training data.
\newblock In \emph{arXiv preprint arXiv: 2205.11482}.

\bibitem[{Ba{\~n}{\'o}n et~al.(2020)Ba{\~n}{\'o}n, Chen, Haddow, Heafield,
  Hoang, Espl{\`a}-Gomis, Forcada, Kamran, Kirefu, Koehn
  et~al.}]{banon2020paracrawl}
Marta Ba{\~n}{\'o}n, Pinzhen Chen, Barry Haddow, Kenneth Heafield, Hieu Hoang,
  Miquel Espl{\`a}-Gomis, Mikel~L Forcada, Amir Kamran, Faheem Kirefu, Philipp
  Koehn, et~al. 2020.
\newblock Paracrawl: Web-scale acquisition of parallel corpora.
\newblock In \emph{Proceedings of the 58th Annual Meeting of the Association
  for Computational Linguistics}, pages 4555--4567.

\bibitem[{Barshan et~al.(2020)Barshan, Brunet, and
  Dziugaite}]{barshan2020relatif}
Elnaz Barshan, Marc-Etienne Brunet, and Gintare~Karolina Dziugaite. 2020.
\newblock Relatif: Identifying explanatory training samples via relative
  influence.
\newblock In \emph{International Conference on Artificial Intelligence and
  Statistics}, pages 1899--1909. PMLR.

\bibitem[{Caswell et~al.(2020)Caswell, Breiner, van Esch, and
  Bapna}]{caswell2020language}
Isaac Caswell, Theresa Breiner, Daan van Esch, and Ankur Bapna. 2020.
\newblock Language id in the wild: Unexpected challenges on the path to a
  thousand-language web text corpus.
\newblock \emph{arXiv preprint arXiv:2010.14571}.

\bibitem[{Cook and Weisberg(1982)}]{cook1982residuals}
R~Dennis Cook and Sanford Weisberg. 1982.
\newblock \emph{Residuals and influence in regression}.
\newblock New York: Chapman and Hall.

\bibitem[{Edunov et~al.(2018)Edunov, Ott, Auli, and
  Grangier}]{edunov2018understanding}
Sergey Edunov, Myle Ott, Michael Auli, and David Grangier. 2018.
\newblock Understanding back-translation at scale.
\newblock \emph{arXiv preprint arXiv:1808.09381}.

\bibitem[{Guo et~al.(2020)Guo, Rajani, Hase, Bansal, and Xiong}]{guo2020fastif}
Han Guo, Nazneen~Fatema Rajani, Peter Hase, Mohit Bansal, and Caiming Xiong.
  2020.
\newblock Fastif: Scalable influence functions for efficient model
  interpretation and debugging.
\newblock \emph{arXiv preprint arXiv:2012.15781}.

\bibitem[{Hampel(1974)}]{hampel1974influence}
Frank~R Hampel. 1974.
\newblock The influence curve and its role in robust estimation.
\newblock \emph{Journal of the american statistical association},
  69(346):383--393.

\bibitem[{Han and Tsvetkov(2020)}]{han2020fortifying}
Xiaochuang Han and Yulia Tsvetkov. 2020.
\newblock Fortifying toxic speech detectors against veiled toxicity.
\newblock \emph{arXiv preprint arXiv:2010.03154}.

\bibitem[{Han et~al.(2020)Han, Wallace, and Tsvetkov}]{han2020explaining}
Xiaochuang Han, Byron~C Wallace, and Yulia Tsvetkov. 2020.
\newblock Explaining black box predictions and unveiling data artifacts through
  influence functions.
\newblock \emph{arXiv preprint arXiv:2005.06676}.

\bibitem[{He et~al.(2019)He, Gu, Shen, and Ranzato}]{he2019revisiting}
Junxian He, Jiatao Gu, Jiajun Shen, and Marc'Aurelio Ranzato. 2019.
\newblock Revisiting self-training for neural sequence generation.
\newblock \emph{arXiv preprint arXiv:1909.13788}.

\bibitem[{Junczys-Dowmunt(2018)}]{junczys2018dual}
Marcin Junczys-Dowmunt. 2018.
\newblock Dual conditional cross-entropy filtering of noisy parallel corpora.
\newblock \emph{arXiv preprint arXiv:1809.00197}.

\bibitem[{Khayrallah and Koehn(2018)}]{khayrallah2018impact}
Huda Khayrallah and Philipp Koehn. 2018.
\newblock On the impact of various types of noise on neural machine
  translation.
\newblock \emph{arXiv preprint arXiv:1805.12282}.

\bibitem[{Koh and Liang(2017)}]{koh2017understanding}
Pang~Wei Koh and Percy Liang. 2017.
\newblock Understanding black-box predictions via influence functions.
\newblock In \emph{International conference on machine learning}, pages
  1885--1894. PMLR.

\bibitem[{Pruthi et~al.(2020)Pruthi, Liu, Kale, and
  Sundararajan}]{pruthi2020estimating}
Garima Pruthi, Frederick Liu, Satyen Kale, and Mukund Sundararajan. 2020.
\newblock Estimating training data influence by tracing gradient descent.
\newblock \emph{Advances in Neural Information Processing Systems},
  33:19920--19930.

\bibitem[{Ram{\'\i}rez-S{\'a}nchez et~al.(2020)Ram{\'\i}rez-S{\'a}nchez,
  Zaragoza-Bernabeu, Ba{\~n}{\'o}n, and Ortiz-Rojas}]{ramirez2020bifixer}
Gema Ram{\'\i}rez-S{\'a}nchez, Jaume Zaragoza-Bernabeu, Marta Ba{\~n}{\'o}n,
  and Sergio Ortiz-Rojas. 2020.
\newblock Bifixer and bicleaner: two open-source tools to clean your parallel
  data.
\newblock In \emph{Proceedings of the 22nd Annual Conference of the European
  Association for Machine Translation}, pages 291--298.

\bibitem[{Schioppa et~al.(2021)Schioppa, Zablotskaia, Vilar, and
  Sokolov}]{schioppa2021scaling}
Andrea Schioppa, Polina Zablotskaia, David Vilar, and Artem Sokolov. 2021.
\newblock Scaling up influence functions.
\newblock \emph{arXiv preprint arXiv:2112.03052}.

\bibitem[{Sennrich et~al.(2015)Sennrich, Haddow, and
  Birch}]{sennrich2015improving}
Rico Sennrich, Barry Haddow, and Alexandra Birch. 2015.
\newblock Improving neural machine translation models with monolingual data.
\newblock \emph{arXiv preprint arXiv:1511.06709}.

\bibitem[{Vaswani et~al.(2017)Vaswani, Shazeer, Parmar, Uszkoreit, Jones,
  Gomez, Kaiser, and Polosukhin}]{vaswani2017attention}
Ashish Vaswani, Noam Shazeer, Niki Parmar, Jakob Uszkoreit, Llion Jones,
  Aidan~N Gomez, {\L}ukasz Kaiser, and Illia Polosukhin. 2017.
\newblock Attention is all you need.
\newblock \emph{Advances in neural information processing systems}, 30.

\bibitem[{Wang et~al.(2018)Wang, Watanabe, Hughes, Nakagawa, and
  Chelba}]{wang2018denoising}
Wei Wang, Taro Watanabe, Macduff Hughes, Tetsuji Nakagawa, and Ciprian Chelba.
  2018.
\newblock Denoising neural machine translation training with trusted data and
  online data selection.
\newblock \emph{arXiv preprint arXiv:1809.00068}.

\end{thebibliography}
\bibliographystyle{acl_natbib}

\vfill
\pagebreak
\onecolumn

\appendix
\section{Appendix}
\setcounter{table}{0}
\renewcommand{\thetable}{A\arabic{table}}
\setcounter{figure}{0}
\renewcommand{\thefigure}{A\arabic{figure}}

\begin{table*}[ht]
\resizebox{\textwidth}{!}
{
\begin{tabular}{cl|cc|ccccc}
\toprule
& Samples & $\nabla_{Full}$ & $\nabla_{Emb}$ & $\nabla_{srcEmb}$ & $\nabla_{trgEmb}$ & $\nabla_{output}$ & $\nabla_{concat}$ \\
\midrule
Probing & Selbst die britische Queen hat ihn schon geadelt. & 1 & 1 & 1 & 1 & 1 & 1 \\
  & Even the British Queen has bestowed an & & & & & \\
  & honour upon him. & & & & & \\
\midrule
1  & \textcolor{red}{Nur fehlten die Beweise.} & 0.358 & 0.284 & 0.024 & 0.225 & 0.401 & 0.319 \\
  & Even the British Queen has bestowed & & & & & \\
  & an honour upon him. & & & & & \\
\midrule
2  & \textcolor{red}{Biologen haben in Hannover untersucht,} & 0.275 & 0.168 & 0.004 & 0.219 & 0.280 & 0.200 \\
  & \textcolor{red}{welchen Effekt das Rufen von Katzenbabys} & & & & &  \\
  & \textcolor{red}{auf erwachsene Tiere hat.} & & & & &  \\
  & Even the British Queen has bestowed & & & & & \\
  & an honour upon him. & & & & & \\
\midrule
3  & Selbst die britische Queen hat ihn schon geadelt. & -0.035 & -0.038 & -0.125 & 0.025 & -0.043 & -0.036 \\
  & \textcolor{red}{The German branch of the Gülen movement} & & & & &  \\
  & \textcolor{red}{also fears that many Turks will flee abroad.} & & & & & \\
\midrule
4  & Selbst die britische Queen hat ihn schon geadelt. & -0.039 & -0.013 & -0.141 & 0.039 & 0.001 & -0.003  \\
  & \textcolor{red}{Demonstrators demanding political change} & & & & & \\
  & \textcolor{red}{in Ethiopia have been met with violent resistance} & & & & & \\
  & \textcolor{red}{by the government.} & & & & & \\
\midrule
5  & Selbst die britische Queen hat ihn schon geadelt. & 0.962 & 0.924 & 0.992 & 0.981 & 0.905 & 0.924  \\
  & Even the British Queen has bestowed & & & & & \\ 
  & an honour upon him\textcolor{red}{!} & & & & & \\
\midrule
6  & Selbst die britische Queen hat ihn schon geadelt\textcolor{red}{!} & 0.908 & 0.899 & 0.912 & 0.949 & 0.935 & 0.935 \\
  & Even the British Queen has bestowed & & & & & \\
  & an honour upon him. & & & & & \\
\bottomrule
\end{tabular}
}
\caption{Another example showing the changes of gradient similarity by selected network components. Segments that are marked in red are perturbed from the probing example. The notation $\nabla_{X}$ indicates the network components used in computing the gradient similarity. $\nabla_{srcEmb}$ has a mean magnitude of 0.051 and 0.007 on random target and random source respectively whereas $\nabla_{output}$ has respectively a mean magnitude of 0.0145 and 0.350. This shows that $\nabla_{output}$ has a tendency of scoring sentence-pairs containing random source higher.}
\label{tab:sensitivity_network_2}
\end{table*}

\begin{table*}[ht]
\resizebox{\textwidth}{!}
{
\begin{tabular}{lll}
\toprule
probing & 1 & Der Film läuft bei uns ab dem 25. \underline{August}.\\
& & The film will be filmed here on 25 \underline{January}. \\
\midrule
$\nabla_{srcEmb}$ & 1 & Die Vereinbarung läuft am 31. \underline{Januar} ab. \\
& & This agreement formally expires on 31 \underline{January}. \\
& 2 & Dieses Gesetz wurde im \underline{August} unterzeichnet. \\
& & It was signed in \underline{January}. \\
& 3 & Die Vereinigten Staaten haben diese Garantie am 15. \underline{August} 1971 aufgegeben. \\
& & The United States abandoned that guarantee on 15 \underline{January} 1971. \\
\midrule
$\nabla_{output}$ & 1 & Der Cardiff-Bericht erscheint Mitte \underline{Januar}. \\
& & The Cardiff report will be published in mid-\underline{January}. \\
& 2 & Eine zweite Tagung ist für \underline{Januar} 2004 vorgesehen. \\
& & A second meeting will be held in \underline{January} 2004. \\
& 3 & Ich hoffe, dass die Dynamik beibehalten und das Siebte Rahmenprogramm \\
& & am 1. \underline{Januar} 2007 auf den Weg gebracht wird. \\
& & I hope that the momentum will be maintained and the Seventh Framework Programme \\
& &  will be launched on 1 \underline{January} 2007. \\
\midrule
probing & 2 & Auch in \underline{Deutschland} finde eine "Hexenjagd" gegen Erdogan-Kritiker statt. \\
& & A 'witch hunt' against Erdogan critics is also taking place in \underline{Italy}. \\
\midrule
$\nabla_{srcEmb}$ & 1 & \underline{Deutschland} ist dagegen.\\
& & \underline{Italy} is opposed to this. \\
& 2 & Dies wäre ein besseres Wirtschaftsmodell für \underline{Deutschland}.\\
& & This would be a better economic model for \underline{Italy}.\\
& 3 & \underline{Deutschland} und China können mehr tun als andere. \\
& & \underline{Italy} and China can do more than others. \\
\midrule
$\nabla_{output}$ & 1 & Eine weitere Lehre für Sarkozy aus \underline{Deutschland} ist, dass ein aufgeklärter  \\
& & korporatistischer Staat unterstützender politischer Führung \\
& & ebenso bedarf wie entgegenkommender Gewerkschaften. \\
& & A further lesson for Sarkozy from \underline{Italy} is that an enlightened corporate state \\
& & needs supportive political leadership as well as accommodating trade unions. \\
& 2 & Insgesamt wurden fast 2 300 Tonnen möglicherweise kontaminiertes Futtermittelfett \\
& & an 25 Futtermittelhersteller in \underline{Deutschland} geliefert. \\
& & A total of almost 2 300 tonnes of potentially contaminated feed fat was delivered\\
& &  to 25 feed manufacturers in \underline{Italy}.  \\
& 3 & Leider Gottes ist der Titel der heutigen Debatte \underline{Italien}.\\
& & Alas, the title of today's debate is \underline{Italy}. \\
\bottomrule
\end{tabular}
}
\caption{Two probing examples with source-hypothesis as input and their top-3 positively influential training instances. $\nabla_{output}$ has a tendency to assign higher scores to sentence-pairs which target side has overlapped tokens but ignoring the similarity of the source side. For example, the pattern “\textit{Januar} -> \textit{January}” occurs more frequently in the ranking than “\textit{August} -> \textit{January}” in probing 1.}
\label{tab:qualitative_analysis}
\end{table*}

\begin{table*}[ht]
\footnotesize
{
\begin{subtable}[h]{\textwidth}
\centering
\begin{tabular}{lccccccc}
\toprule
\multirow{2}{*}{$\nabla$(Probing)} & \multirow{2}{*}{+/-} & \multicolumn{3}{c}{Precision} \\
  & & $\nabla_{srcEmb}$ & $\nabla_{encoder}$ & $\nabla_{trgEmb}$ & $\nabla_{output}$ & $\nabla_{concat}$ & $\nabla_{Full}$ \\
\midrule
$\nabla$(HYP) & + & 0.846 & 0.485 &	0.334 &	0.720 &	0.722 &	0.503 \\
$\nabla$(REF) & - & 0.876 &	0.432 &	0.303 &	0.794 &	0.805 &	0.481 \\
$\nabla$(CorrHYP) & - & 0.930 &	0.494 &	0.324 &	0.905 &	0.919 &	0.531 \\ [2mm]
$\nabla$(HYP\textsubscript{Mask}) & + & 0.893 &	0.581 &	0.347 &	0.840 &	0.844 &	0.654 \\
$\nabla$(HYP\textsubscript{MaskExact}) & + & 0.957 & 0.862 & \bf 0.474 & 0.910 & 0.916 & 0.862 \\
$\nabla$(CorrHYP\textsubscript{MaskExact})& - & 0.989 &	0.903 &	0.467 &	\bf 0.992 &	\bf 0.994 &	0.924 \\ [2mm]
$\nabla$(HYP) - $\nabla$(REF) & + & 0.930 & 0.523 &	0.321 &	0.856 &	0.855 &	0.584 \\
$\nabla$(HYP) - $\nabla$(CorrHYP) & + & \bf 1.000 & \bf	0.985 &	0.458 &	0.971 &	0.980 &	\bf 0.987 \\
\bottomrule
\end{tabular}
\caption{Retrieval performance for top-1\% influential training examples}
\end{subtable}
\vspace*{5mm}
\newline
\begin{subtable}[h]{\textwidth}
\centering
\begin{tabular}{lccccccc}
\toprule
\multirow{2}{*}{$\nabla$(Probing)} & \multirow{2}{*}{+/-} & \multicolumn{3}{c}{Precision} \\
  & & $\nabla_{srcEmb}$ & $\nabla_{encoder}$ & $\nabla_{trgEmb}$ & $\nabla_{output}$ & $\nabla_{concat}$ & $\nabla_{Full}$ \\
\midrule
$\nabla$(HYP) & + & 0.765 &	0.399 &	0.301 &	0.644 &	0.646 & 0.442 \\
$\nabla$(REF) & - & 0.799 &	0.382 &	0.297 &	0.693 &	0.700 &	0.437 \\
$\nabla$(CorrHYP) & - & 0.844 &	0.402 &	0.299 &	0.781 &	0.789 &	0.455 \\ [2mm]
$\nabla$(HYP\textsubscript{Mask}) & + & 0.848 &	0.478 &	0.311 &	0.829 &	0.831 &	0.567 \\
$\nabla$(HYP\textsubscript{MaskExact})  & + & 0.936 & 0.794 & \bf 0.380 & 0.904	& 0.908	& 0.825 \\ 
$\nabla$(CorrHYP\textsubscript{MaskExact}) & - & 0.962 & 0.821 & 0.372 & \bf 0.958 & \bf 0.960 & 0.875 \\ [2mm]
$\nabla$(HYP) - $\nabla$(REF) & + & 0.855 & 0.442 &	0.307 &	0.764 &	0.765 &	0.515 \\
$\nabla$(HYP) - $\nabla$(CorrHYP) & + & \bf 0.986 & \bf	0.884 &	0.371 &	0.935 &	0.939 & \bf	0.931 \\
\bottomrule
\end{tabular}
\caption{Retrieval performance for top-10\% influential training examples}
\end{subtable}
}
\caption{Retrieval performance measured in (macro) averaged precision over all error patterns (extended version of Table~\ref{tab:avgPrecision}). $\nabla(Probing)$ refers to the gradient with input ‘source-Probing’. HYP, REF and CorrHYP stands for hypothesis, reference and corrected hypothesis respectively. ``+'' (``-'') indicates that positively (negatively) influential training instances were retrieved. $\nabla_{X}$ indicates network components used in computing the gradient, $\nabla_{concat}$ indicates concatenation of $\nabla_{srcEmb}$, $\nabla_{trgEmb}$ and $\nabla_{output}$. We mark the best result per column in bold.}
\label{tab:avgPrecisionFull}
\end{table*}

\begin{figure*}[h]
\begin{subfigure}[t]{0.45\textwidth}
\includegraphics[width=\textwidth]{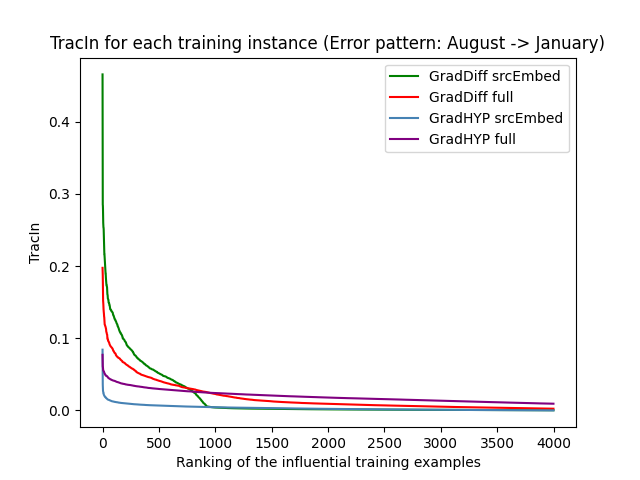}
\end{subfigure}\hspace{\fill}
\begin{subfigure}[t]{0.45\textwidth}
\includegraphics[width=\textwidth]{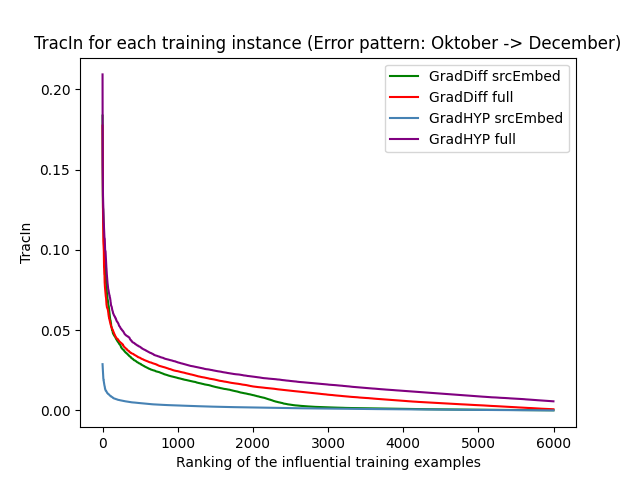}
\end{subfigure}

\begin{subfigure}[t]{0.45\textwidth}
\includegraphics[width=\textwidth]{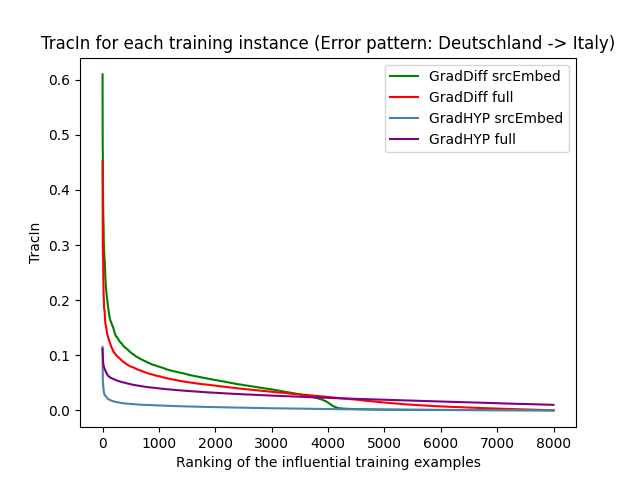}
\end{subfigure}\hspace{\fill}
\begin{subfigure}[t]{0.45\textwidth}
\includegraphics[width=\textwidth]{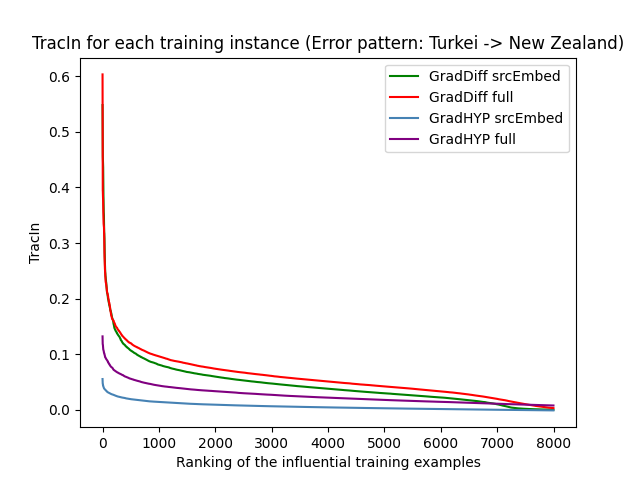}
\end{subfigure}
\caption{TracIn of the top-50\% positively influential training examples. In each subfigure, we randomly select a probing example from each error pattern to compute its influence using gradient difference w.r.t. 1) source embedding (GradDiff srcEmbed), and 2) entire model (GradDiff full) as well as using vanilla-IF with source-hypothesis as input w.r.t. 1) source embedding (GradHYP srcEmbed), and 2) entire model (GradHYP full).}
\label{fig:IF_ranking_half}
\end{figure*}

\begin{table*}[ht]
\resizebox{\textwidth}{!}
{
\begin{tabular}{lll}
\toprule
probing & 1 & Golfer Langer erhält die Sportpyramide \\
& & Golfer Langer erhält die Sportpyramide \\
\midrule
$\nabla_{srcEmb}$ & 1 & Binnenmarktanzeiger \\
& & Binnenmarktanzeiger \\
& 2 & Vollständige Liste der ausgewählten Aussteller: \\
& & Vollständige Liste der ausgewählten Aussteller: \\
& 3 & Dimiter TZANTCHEV Ständiger Vertreter \\
& & Dimiter TZANTCHEV Ständiger Vertreter \\
\midrule
$\nabla_{Full}$ & 1 & Erstellung einzelstaatlicher Aktionspläne für die Verhütung von Verletzungen durch die Mitgliedstaaten. \\
& & Erstellung einzelstaatlicher Aktionspläne für die Verhütung von Verletzungen durch die Mitgliedstaaten. \\
& 2 & Für weitere Informationen wenden Sie sich bitte an die Dienststelle Außenbeziehungen Europäischer Rechnungshof \\
& & Für weitere Informationen wenden Sie sich bitte an die Dienststelle Außenbeziehungen Europäischer Rechnungshof \\
& 3 & Dimiter TZANTCHEV Ständiger Vertreter \\
& & Dimiter TZANTCHEV Ständiger Vertreter \\
\midrule
probing & 2 & Die demokratische Bewerberin kündigt gar die größte Investition in neue Arbeitsplätze seit dem Zweiten Weltkrieg an. \\
& & Die demokratische Bewerberin kündigt gar die größte Investition in neue Arbeitsplätze seit dem Zweiten Weltkrieg an. \\
\midrule
$\nabla_{srcEmb}$ & 1 & Die Krise hat die großen Unterschiede innerhalb der EU deutlich gemacht.\\
& & Die Krise hat die großen Unterschiede innerhalb der EU deutlich gemacht. \\
& 2 & Die Regierungskonferenz ist nur eine Versammlung aller Regierungen.\\
& & Die Regierungskonferenz ist nur eine Versammlung aller Regierungen.\\
& 3 & Die Entschließung wird uns dabei helfen, auf einer soliden Grundlage in die nächste Phase der Entwicklung \\
& & einer Meeresstrategie einzutreten. \\
& & Die Entschließung wird uns dabei helfen, auf einer soliden Grundlage in die nächste Phase der Entwicklung \\
& & einer Meeresstrategie einzutreten. \\
\midrule
$\nabla_{Full}$ & 1 & Die Partei für Freiheit möchte dafür sorgen, dass die niederländische Öffentlichkeit nicht länger als \\
& & Geldautomat Europas behandelt wird.  \\
& & Die Partei für Freiheit möchte dafür sorgen, dass die niederländische Öffentlichkeit nicht länger als \\
& & Geldautomat Europas behandelt wird. \\
& 2 & Die russische Regierung hat geschätzt, dass ein Drittel aller Wasserleitungen dringend ersetzt werden muss. \\
& & Die russische Regierung hat geschätzt, dass ein Drittel aller Wasserleitungen dringend ersetzt werden muss. \\
& 3 & Die internationale Gemeinschaft erkannte ihn einstimmig an. \\
& & Die internationale Gemeinschaft erkannte ihn einstimmig an. \\
\bottomrule
\end{tabular}
}
\caption{Two probing examples with copied training instances as input and their top-3 positively influential training instances. Both $\nabla_{srcEmb}$ and $\nabla_{Full}$ can retrieve copied instances in the training subset given a probing instance of copied source sentence which is lexically different.}
\label{tab:qualitative_analysis_untranslatedSRC}
\end{table*}

\end{document}